# A Multi-Scale and Multi-Depth Convolutional Neural Network for Remote Sensing Imagery Pan-Sharpening

Qiangqiang Yuan, *Member, IEEE*, Yancong Wei, *Student Member, IEEE*, Xiangchao Meng, *Student Member, IEEE*, Huanfeng Shen, *Senior Member, IEEE*, Liangpei Zhang, *Senior Member, IEEE*,

*Abstract*--Pan-sharpening is a fundamental and significant task in the field of remote sensing imagery processing, in which high-resolution spatial details from panchromatic images are employed to enhance the spatial resolution of multi-spectral (MS) images. As the transformation from low spatial resolution MS image to high-resolution MS image is complex and highly non-linear, inspired by the powerful representation for non-linear relationships of deep neural networks, we introduce multi-scale feature extraction and residual learning into the basic convolutional neural network (CNN) architecture and propose the multi-scale and multi-depth convolutional neural network (MSDCNN) for the pan-sharpening of remote sensing imagery. Both the quantitative assessment results and the visual assessment confirm that the proposed network yields high-resolution MS images that are superior to the images produced by the compared state-of-the-art methods.

## I. INTRODUCTION

In remote sensing images, panchromatic (PAN) images have a very high spatial resolution with the cost of lacking spectral band diversities, while multi-spectral (MS) images contain rich spectral information but have a lower spatial resolution. However, due to the technical limitations of sensors and other factors, remote sensing images with both high spatial and spectral resolutions, which are highly desirable in many remote sensing applications, are currently unavailable. Therefore, researchers have made efforts to fuse PAN images with MS images to produce an image with both high spatial and spectral resolutions, which is a process that is also called "pan-sharpening".

To date, a variety of pan-sharpening methods have been proposed, and most of them can be divided into there major categories: **1) Component substitution (CS)-based methods.** This type of method traditionally transform the MS image into a suitable domain. The specific component representing the spatial information of the MS image is then replaced by the PAN image, and inverse transformation is performed to reconstruct the fused image. Examples of CS-based methods are the typical intensity-hue-saturation (IHS) fusion methods [1][2], the principal component analysis (PCA) fusion method[3], the Gram-Schmidt (GS) fusion method [4], and adaptive component-substitution-based satellite image fusion using partial replacement [5]. It should be noted that, in this group of methods, analysis of the correlation between the replaced MS component and the PAN image has a great influence on the fusion result. **2) Multiresolution analysis (MRA)-based methods.** Compared with the traditional CS-based methods, the MRA-based methods generally have better spectral information preservation. In general, this type of method first extract the spatial structures from the PAN image by wavelet transform, Laplacian pyramid, etc., and then the extracted spatial structure information is injected into the up-sampled MS images to obtain the fused image. Examples of this type of method are the fusion methods based on wavelet transform [6] or curvelet transform [7], the analysis of modulation transfer function (MTF) [8][9], and the Smoothing Filter based Intensity Modulation (SFIM) method [10]. A combination of CS and MRA has also been recently proposed to enhance the spatial-spectral unified fidelity of fused images [11]. However, these types of methods generally produce spatial distortion, and there is a strict requirement for accurate co-registration between the PAN and up-sampled MS images. **3) Model-based optimization (MBO) approaches.** These types of methods are based on the image observation models and regard the solution of the fused image as an ill-posed inverse problem. Generally, the fusion images can be solved by minimizing a loss function with the prior constraints, from minimum mean square error (MMSE) that has been employed to form the the Band-Dependent Spatial Detail (BDSD) model [12], non-local optimization based on *k*-means clustering [13], to advanced regularization operators such as Bayesian posterior probability [14], adaptive regularization based on normalized Gaussian distribution [15], Total Variation (TV) operators [16][17], and Sparse Reconstruction (SR)-based fusion methods [18]. Among them the SR-based fusion methods are among the most advanced algorithms for general signal processing tasks. The basic idea of the SR-based methods is that the low spatial resolution MS images and the high spatial resolution PAN images are decomposed into differently scaled dictionaries (high spatial resolution dictionary and low spatial resolution dictionary) and some sparse coefficient, and the latter can be shared during the reconstruction at the target resolution level to obtain the fused image.

Although a variety of pan-sharpening methods have been proposed, the disadvantages of these four major types of methods are hard to ignore. In the CS- and MRA-based fusion methods, the transformation from observed images to fusion targets is not rigorously modeled and distortion in the spectral domain is very common. In the results of the MBO-based methods, the spectral distortion can be reduced by better modeling of the transformation, and a much higher accuracy can be produced, but the linear simulation from the observed and fusion image is still a limitation, especially when the spectral coverages of the PAN and MS images do not fully overlap and lead to the fusion process being highly non-linear. Furthermore, in the MBO-based methods, the design of the optimal fusion energy function is heavily reliant on prior knowledge, and on images with different distributions and quality degeneration, these models are not robust. Furthermore, solving the regularization models generally requires iterative computing, which is time-consuming and may cause incidental errors, especially for the images with a large size.

To overcome those shortcomings, advanced algorithms have been introduced in recent years, and among them, the deep learning models are some of the most promising approaches. Deep learning models are built with multiple transforming layers, and in each layer, its input is linearly filtered to produce an output, and multiple layers are stacked to form a total transformation with high non-linearity. The most outstanding advantage of the deep learning models is that all the parameters included in the model can be updated under the supervision of training samples, and thus the requirement for prior knowledge is reduced and much higher fitting accuracies can be expected.

For both natural images and remote sensing images, in the field of most low-level vision tasks, e.g., image denoising, deblurring, super-resolution, inpainting, etc. [23]–[31], deep learning based methods have achieved state-of-the-art accuracies in recent years, and their performances are continuously being improved. However, in the field of pan-sharpening, only limited studies have been undertaken in recent years to introduce deep learning models. Examples are the sparse deep neural network [32] and the pan-sharpening neural network (PNN) [33], the latter of which has achieved impressive performance gains. However, as the design of the PNN is completely borrowed from the super-resolution CNN (SRCNN) proposed in [22], which is considered a relatively simple and shallow architecture when compared to its later derivations [23][27][28][30], there is still plenty of room for improvement. To exploit the advantages of deep learning and overcome the shortcomings of the current methods, we propose an original network that is specifically designed for the pan-sharpening task, while it can also be generalized for other types of image restoration problems. The framework consists of a PNN and a deeper multi-scale neural network (MSNN). The former network performs simple feature extraction, while the latter network contains multi-scale feature extraction layers and builds a deep architecture. We believe that as the scale of features greatly varies among different ground objects from multiple sensors, introducing multi-scale feature extraction can help to learn more robust convolutional filters, and thus the fusion accuracy can be advanced from the current state-of-the-art level. This assumption is fully supported by the experimental results, which are described in Section IV.

The rest of this paper is organized as follows. The background knowledge to pan-sharpening and the related deep learning works are introduced in Section II. The detailed architecture of the proposed multi-scale and multi-depth convolutional neural network (MSDCNN) is described in Section III. The results of the pan-sharpening accuracy assessment are presented in Section IV. Finally, a discussion and the conclusion are given in Section V.

## II. BACKGROUND

### A. Pan-Sharpening Based on Linear Models

Assuming that the low-resolution MS image is considered as a degraded observation $g_{MS}$, then the PAN image $g_{PAN}$ that matches $g_{MS}$ is included to guide the prediction process of the high-resolution spatial details in the ground truth $f_{MS}$. The main aim of the pan-sharpening task is to preserve the unified spatial-spectral fidelity for the fused image. For a low-resolution MS image $g_{MS}$ with $S$ spectral bands, we denote the pan-sharpened result as $F_{MS}$, which is an estimation of $f_{MS}$, and then the constraint function of the MS image pan-sharpening can be formed as:

$$\arg\min_{F_{MS}} \sum_{i=1}^{S} \left\| f_{MS(i)} - F_{MS(i)} \right\|_2^2 \quad (1)$$

where $F_{MS}$ is obtained from a fusion function:

$$F_{MS} = P(g_{MS}, g_{PAN}) \quad (2)$$

In (2), $P(.)$ represents the pan-sharpening process. In the traditional MBO approaches, both $g_{MS}$ and $g_{PAN}$ are considered as degraded observations of $f_{MS}$ in relative domains, and the fusion process is simulated under a linear framework as:

$$\begin{bmatrix} g_{MS} \\ g_{PAN} \end{bmatrix} = \begin{bmatrix} DHf_{MS} \\ Rf_{MS} \end{bmatrix} + \begin{bmatrix} N_{MS} \\ N_{PAN} \end{bmatrix} \quad (3)$$

where $D$ is a down-sampling matrix in the spatial domain, and similarly, $R$ is the spectral response matrix of the PAN channel of the sensor, which down-samples the latent ground truth along the spectrum. $H$ is a blurring matrix, while $N_{MS}$ and $N_{PAN}$ are the additive noise, which is assumed to be Gaussian distributed. Therefore, (2) is linearly fitted by solving an optimization function as:

$$\arg\min_{F_{MS}} \{ \lambda_1 \| DHF_{MS} - g_{MS} \|_P^2 + \lambda_2 \| RF_{MS} - g_{PAN} \|_P^2 + \lambda_3 \varphi(F_{MS}) \} \quad (4)$$

in which $\lambda_i (i=1,2,3)$ represents the weights that control the contributions of the three items, and the constraint operator $\varphi(F_{MS})$ is based on reasonable assumptions and prior knowledge to reduce the ill-posed property of the problem.

However, it should be noted that in the pan-sharpening process, the bandwidths of the PAN and MS images are not guaranteed to fully overlap. For example, the MS bandwidth of WorldView-2 ranges from 400 nm to 1040 nm and is divided into eight bands, and its PAN bandwidth covers 450–800 nm. Thus, if we keep simulating the transformation $P(.)$ from a linear perspective, as in (4), it is difficult to merge the down-sampled spectra of the PAN images into the spectra of the MS images while preserving the fidelity of the latter. The drawbacks of such linear models can be explained as follows. Firstly, a satisfactory accuracy can rarely be achieved when linear functions are employed to fit complex transformations, especially for ill-posed inverse problems. Secondly, prior knowledge that has been artificially introduced into the problem, e.g., the design of $\varphi(F_{MS})$, is not guaranteed to be suitable for general tasks and may increase the system error. Furthermore, for images of many complex circumstances and from different sensors, the value of $\lambda_i$ needs to be empirically chosen and lacks a robust solution. Thus, the abilities of the linear optimization models are somewhat limited.

To overcome the drawbacks of the linear models, a non-linear function is needed to fit the fusion process, which requires us to employ a different point of view to investigate the correlation between $g_{MS}$, $g_{PAN}$, and $f_{MS}$. Therefore, the idea of deep learning is adopted, and is introduced in the next sub-section.

*B. Deep Learning for Pan-Sharpening*

As illustrated in Fig. 1, for the texture details contained in $g_{PAN}$, we regard them as high-frequency components of $f_{MS}$, and the coarse spatial structures of $G_{MS}$ are regarded as low-frequency components. Thus, we can employ a filtering function to extract the features $f_{lowfreq}$ and $f_{highfreq}$, and merge them to yield the high-resolution estimation $F_{MS}$.

How do we obtain a set of filters that can accurately extract complex features from various ground scenes, without causing spectral distortion? The recently developed deep learning approach is one of the most advanced answers to this problem. In the different deep learning networks, convolutional neural networks (CNNs) are a branch of the deep learning models that has impressively swept the field of computer vision and image processing in recent years. In this paper, it is introduced as a prototype of our proposed methodology. Compared to the traditional hand-crafted extractors for features, the superiority of CNNs can be explained with two concepts—"deep" and "learning"— which are explained in the following:

**Deep**: The architectures of CNNs are formed by stacking multiple convolutional layers. Although each of these layers functions as a linear filtering process, a whole network is able to fit a very complex non-linear transformation that maps $\{G_{MS}, g_{PAN}\}$ to $f_{MS}$. The non-linearity and fitting ability of CNNs are not limited to a certain level, as the depth of the network can be infinitely expanded along the direction in which the layers are stacked.

**Learning:** To extract features from $G_{MS}$ and $g_{PAN}$, the filtering process in every convolutional layer of a CNN is executed using convolutional kernels. With the supervision of $f_{MS}$ as a target, the network iteratively updates all the kernels to seek an optimal allocation, and thus it is defined as a **"learning"** process. When the loss between $f_{MS}$ and $F_{MS}$ reaches a satisfactory convergence, the learning of the network is finished and an accurate end-to-end function is obtained for the pan-sharpening. The flowchart of training a deep CNN on a training dataset is shown in Fig. 2.

**Pan-sharpening with a basic CNN:** As mentioned above, $G_{MS}$ and $g_{PAN}$ are fed into a CNN to directly yield a fused image $F_{MS}$. In the network, the input images are passed through $L$ layers, and the filtering process executed in the $n$-th layer can be described as:

$$F_n = P_n(F_{n-1}) \qquad (5)$$

where $F_n$ is the output of the $n$-th layer. Thus, the fusion process can be described as follows:

$$F_0 = G = \{G_{MS}, g_{PAN}\}, \text{Size}: H \times W \times (S+1) \qquad (6)$$

$$F_n = P_n(F_{n-1}) = ReLU(W_n \circ F_{n-1} + b_n), \\ \text{Size}: H \times W \times C_n, n=1,\ldots,L-1 \qquad (7)$$

$$F_{MS} = F_L = W_L \circ F + b_L, \text{Size}: H \times W \times S \qquad (8)$$

where $\circ$ represents three-dimensional convolution, which is the feature extractor in $P_n(F_{n-1})$, and $W_n$ contains $C_n$ groups of convolutional kernels, where the size of each group is $h_n \times w_n \times C_{n-1}$, and $b_n$ is a bias vector with the size of $1 \times 1 \times C_n$. Thus, for the $n$-th layer, $C_n$ represents the spectral dimensionality of its output and can be artificially set. The rectified linear unit (ReLU) is used to introduce non-linearity in the function:

$$ReLU(x) = \max(x, 0) \qquad (9)$$

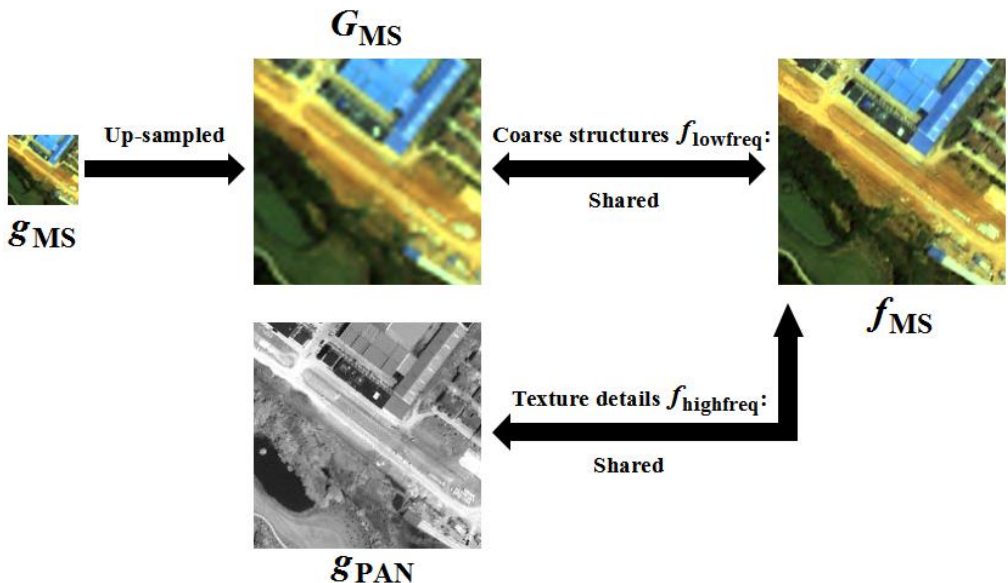

Fig. 1. Visual correlation between a low-resolution MS image, a PAN image, and a high-resolution MS image.

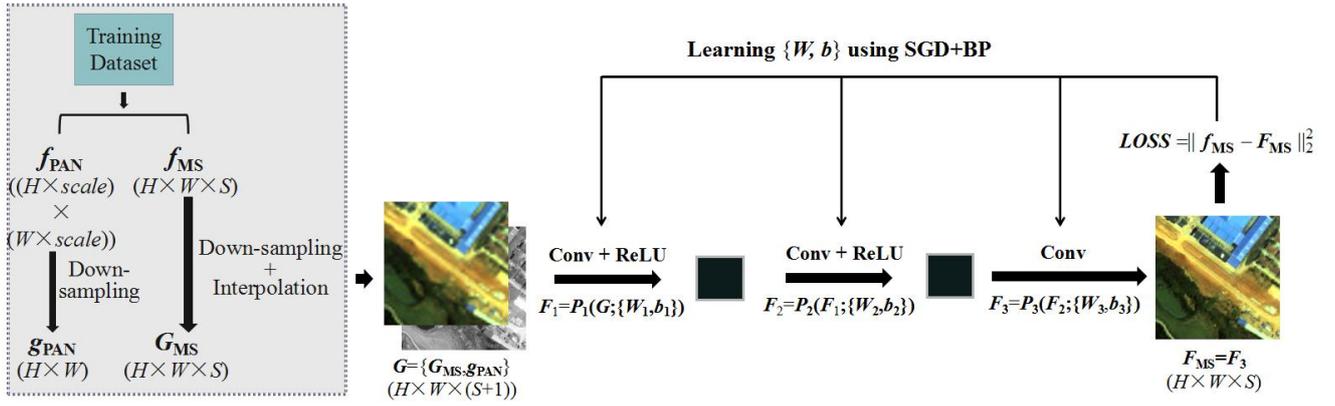

Fig. 2. Flowchart of basic CNN-based pan-sharpening.

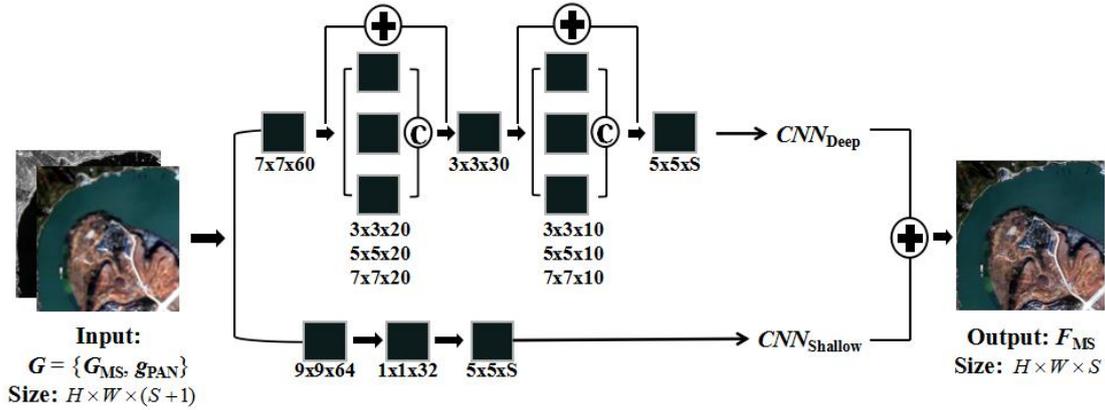

Fig 3. Flowchart of passing the input MS image through MSDCNN to yield a fused result.

## III. PROPOSED NETWORK: MSDCNN

Based on the basic architecture of a CNN with three convolutional layers for pan-sharpening, as previously mentioned, we introduce two concepts to improve the architecture of the network: the **multi-scale feature extraction block** and **skip connection.** The proposed MSDCNN contains two sub-networks: A fundamental three-layer CNN with the same architecture as in [22] and [33], and a deeper CNN with two multi-scale convolutional layer blocks. The whole architecture of MSDCNN is displayed in Fig. 3.

*A. Multi-Scale Feature Extraction Block*

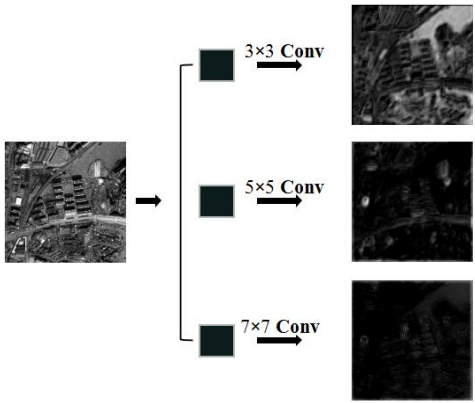

Fig. 4. Feature maps extracted by convolutional filters with three different sizes, which are selected from the first layer of a trained MSDCNN model.

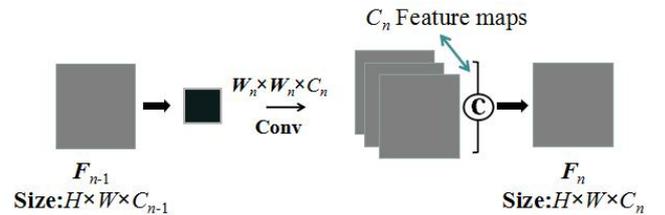

(a) A basic convolutional layer.

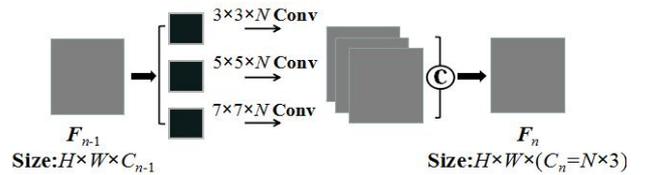

(b) A convolutional layer for multi-scale feature

Fig. 5. The difference between a basic convolutional layer and a layer for multi-scale feature extraction, where **C** stands for concatenating images along the spectral dimension.

As mentioned before, the coarse structures and texture details are the features that need to be extracted from ground objects and scenes. In remote sensing imagery with a meter- or sub-meter-level spatial resolution, the sizes of the ground objects vary from very small neighborhoods to large regions containing thousands of pixels, and a ground scene may cover many objects with various sizes. From the feature maps displayed in Fig. 4, it is indicated that the features with a smaller scale, such as the short edges of buildings and the textures of vegetation, tend to respond to convolutional filters with a smaller size, while the coarse structures tend to be extracted by larger filters.

To make adequate use of the rich spatial information in high-resolution imagery and improve the robustness of the feature extraction among various and complex ground scenes, we introduce the multi-scale convolutional layer block, which was applied to image super-resolution in [30] and classification in [35].

As illustrated in Fig. 5, in the $n$-th layer, three sizes are set for the convolutional kernels contained in the multi-scale layer block: $3 \times 3$, $5 \times 5$, and $7 \times 7$. For each size, $N$ groups of kernels are employed to produce $N$ feature maps, and they are concatenated along the spectral dimension to form the output.

*B. Skip Connection*

As discussed in Section II-B, in CNNs, stacking more layers can lead to higher non-linearity and can help to fit complex transformations more accurately. Visualized feature maps show that when an image is passed through a deeper network, the features extracted from it can be more abstract and representative [36], [37]. However, there is a significant problem in that in the training process of a deep CNN, the gradients of the loss to the network parameters are severely diminished during the back-propagation from output to input. Thus, in layers that are close to the input, updating of the convolutional kernels and bias vectors becomes too slow to reach the optimal allocation of all parameters.

In [22] and [33], it was indicated that for the fundamental architecture of a CNN, $L = 3$ is an upper limit to the depth of the network, and adding more layers can no longer boost the accuracy performance, while the increase in training time also becomes unacceptable. To deal with the problem, residual learning [38] is now considered to be one of the most effective solutions for training deep CNNs, in which the convolutional filtering process $F_n = P_n(F_{n-1})$ is replaced with $F_n = F_{n-1} + P_n(F_{n-1})$, and thus the residual $F_n - F_{n-1}$ becomes the target of the prediction. This simple and effective architecture is called a "skip connection". It is assumed that the distribution of features in the residual image is very sparse and most of the pixel values are close to zero. Thus, the loss- parameters surface of a residual learning function becomes much smoother than the surface of a regular CNN, and the distances from the local minimum points to the optical minimum are shortened.

In [27], an end-to-end skip connection $F = G + P(G)$ was designed to train a very deep CNN for image super-resolution, aiming to use the whole network to directly predict the residual image $f - G$ from the input low-resolution image $G$. However, for the pan-sharpening task, the end-to-end architecture is not suitable due to the different sizes of $G = \{G_{MS}, g_{PAN}\}$ (size: $H \times W \times (S+1)$) and $f_{MS}$ (size: $H \times W \times S$). Thus, in the proposed network, a connection that only skips one layer is set for the block, as illustrated in Fig. 6.

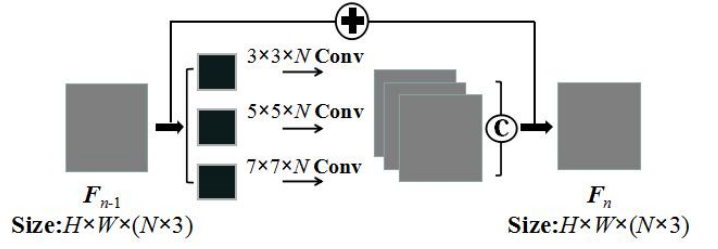

Fig 6. Complete architecture of the proposed multi-scale convolutional layer block with a short-distance skip connection.

*C. Joint Learning for MSDCNN*

As described in Fig. 3, the images output from the two sub-networks of MSDCNN are summed for a final estimation:

$$F_{MS} = MSD(G, \{W, b\}) \\ = CNN_{shallow}(G; \{W_{shallow}, b_{shallow}\}) \\ + CNN_{deep}(G; \{W_{deep}, b_{deep}\}) \quad (10)$$

where all the parameters contained in MSDCNN are jointly learned:

$$\arg\min_{W,b} \sum_{i=1}^{S} \left\| f_{MS(i)} - (MSD(G, W, b))_{(i)} \right\|_2^2 \quad (11)$$

To iteratively learn the optimal allocation of $\{W, b\}$, we let $\{W^t, b^t\}$ represent the values of $\{W, b\}$ in the $t$-th iteration after random initialization, and $F_{MS}^t$ stands for the output from $\{W^t, b^t\}$. The current loss is then:

$$LOSS^t = \sum_{i=1}^{S} \left\| f_{MS(i)}^t - F_{MS(i)}^t \right\|_2^2 \quad (12)$$

By computing the derivatives of $LOSS^t$ to $\{W^t, b^t\}$, the gradients are obtained as:

$$\{\delta W^t, \delta b^t\} = \left\{ \begin{array}{l} \left. \dfrac{\partial LOSS^t(\{W, b\}; G)}{\partial W} \right|_{W=W^t, b=b^t} \\ \left. \dfrac{\partial LOSS^t(\{W, b\}; G)}{\partial b} \right|_{W=W^t, b=b^t} \end{array} \right\} \quad (13)$$

Stochastic gradient descent (SGD) is also applied as an effective way to accelerate the training process. Instead of computing the gradient for a single image, a batch of input images $\{G_1, ..., G_{Batchsize}\}$ are fed into the network in the $t$-th iteration to yield multiple outputs $\{F_{MS_1}^t, ..., F_{MS_{Batchsize}}^t\}$ and an average loss is defined as:

$$\overline{LOSS}^t = \frac{1}{Batchsize} \sum_{b=1}^{Batchsize} \sum_{i=1}^{S} \left\| f_{MS_b(i)}^t - F_{MS_b(i)}^t \right\|_2^2 \quad (14)$$

An input image is then randomly picked from $\{G_1, ..., G_{Batchsize}\}$ and used as $G$ in (10) for computing the gradients. With $\{\delta W^t, \delta b^t\}$ known, $\{W^t, b^t\}$ can be updated using a classic momentum (CM) algorithm [39]. We let $\theta = \{W, b\}$ represent all the parameters in the network, and then the updating of $\theta$ as follows:

$$\Delta \boldsymbol{\theta}^t = \mu \cdot \Delta \boldsymbol{\theta}^{t-1} - \varepsilon \cdot \delta \boldsymbol{\theta}^t \qquad (15)$$

$$\boldsymbol{\theta}^{t+1} = \boldsymbol{\theta}^t + \Delta \boldsymbol{\theta}^t \qquad (16)$$

where $\mu$ is the momentum and $\varepsilon$ is the learning rate. During the training process, gradient clipping is also necessary to avoid gradient explosion. In each iteration, a summed L2-norm of all the gradients is limited, which means that $\delta \boldsymbol{W}^t$ and $\delta \boldsymbol{b}^t$ are clipped as:

$$\left\{ \delta \boldsymbol{W}^t, \delta \boldsymbol{b}^t \right\}_{\text{Clipped}} = \left\{ \frac{\delta \boldsymbol{W}^t}{\left\| \delta \boldsymbol{W}^t \right\|_2^2 / 0.1}, \frac{\delta \boldsymbol{b}^t}{\left\| \delta \boldsymbol{b}^t \right\|_2^2 / 0.1} \right\} \qquad (17)$$

## IV. EXPERIMENTAL RESULTS AND DISCUSSION

### A. Experimental Settings

**1) Datasets:** To simulate the fusion transformation, original MS images with different numbers of spectral bands from QuickBird and WorldView-2 sensors were used as the ground truth $f_{\text{MS}}$, and we then down-sampled $f_{\text{MS}}$ and used bicubic interpolation to obtain the low-resolution MS image $G_{\text{MS}}$. The PAN image was also down-sampled as $g_{\text{PAN}}$, and thus the ratios of the scales among $G_{\text{MS}}$, $g_{\text{PAN}}$, and $f_{\text{MS}}$ were kept the same to the real situation.

For training and simulated testing of the proposed MSDCNN, we collected two large datasets from QuickBird and WorldView-2 images, which were divided into smaller patches to separately train two networks with different numbers of input bands. Details of the datasets used in the experiments are listed in Table I. It should be noted that the number of quantitatively tested samples included in our datasets (two datasets for the quantitative assessment, 240 images in total, with a spatial size of 250×250) was much larger than in the referenced papers; for instance, in [19], three datasets and three images with a spatial size of 600×600 were used, and in [33], three datasets and 150 images with a spatial size of 320×320 were considered.

For the real-data experiments, another smaller dataset was collected from a group of IKONOS images to test the network, and the network was tested on the WorldView-2 dataset with eight bands. The 112 patches in the real-data experiment for images with four bands were collected by fully segmenting the seven scenes of IKONOS images, while the 28 patches in the real-data experiment for images with eight bands were selected from the two test scenes of WorldView-2 images, covering regions of impervious surfaces, water bodies, and urban vegetation.

**2) Model Implementation:** For each dataset, MSDCNN was trained for 300 epochs (about 250,000 iterations), and the batch size was set to 64. To apply CM with SGD, $\mu = 0.9$ and $\varepsilon = 0.1$ were used as the default settings. With the Caffe [40] deep learning framework supported by a GPU (NVIDIA Quadro M4000) and CUDA 7.5, the training process for each model cost roughly eight hours.

TABLE I
DETAILS OF THE THREE DATASETS USED IN THE TRAINING AND TESTING

| Sensor | MS bands | Scenes | Covered regions | Training | Simulated experiments | Real-data experiments |
|---|---|---|---|---|---|---|
| QuickBird | 4 | 4 | Nanchang, China (for training); Shenzhen, China (for training); Wuhan, China (for testing); Yichang, China (for testing) | **Patches**: 51648 **Input size**: $41 \times 41 \times 5$ **Output size**: $41 \times 41 \times 4$ | **Patches**: 160 **Input Size**: $250 \times 250 \times 5$ **Output Size**: $250 \times 250 \times 4$ | Not included |
| IKONOS | 4 | 7 | Wuhan, China (all for testing) | Not included | Not included | **Patches**: 112 **Input size**: $400 \times 400 \times 5$ **Output size**: $400 \times 400 \times 4$ |
| WorldView-2 | 8 | 4 | San Francisco, United States (two scenes for training, two scenes for testing) | **Patches**: 59840 **Input size**: $41 \times 41 \times 9$ **Output size**: $41 \times 41 \times 8$ | **Patches**: 80 **Input size**: $250 \times 250 \times 9$ **Output size**: $250 \times 250 \times 8$ | **Patches**: 28 **Input size**: $800 \times 800 \times 9$ **Output size**: $800 \times 800 \times 8$ |

Testing of all the convolutional networks was performed with the support of MatConvNet [41] on a Dell Tower 7810 workstation with an Intel CPU (Xeon E5-2620 v3 @ 2.40 GHz).

**3) Compared Algorithms:** For the numeric and visual assessment, seven traditional and state-of-the-art algorithms were used, representing different branches of pan-sharpening methods: Gram-Schmidt (GS) [4] and partial replacement adaptive component substitution (PRACS) [5] belonging to CS; the modulation transfer function based generalized Laplacian pyramid (MTF-GLP) [8], smoothing filter based intensity modulation (SFIM) [10], and additive wavelet luminance proportion (AWLP) [42] belonging to multiresolution analysis; two-step sparse coding (TSSC) [19] based on regularization constraint model; and in the deep learning field, the pan-sharpening neural network (PNN) [33] based on a basic CNN containing three layers was considered as the main competitor to the proposed MSDCNN. We also feel thankful to the author of [43] for providing the toolbox that helped us to implement five of the seven referenced algorithms, except TSSC and PNN.

*B. Simulated Experiments*

In these experiments, the PAN and MS images were down-sampled to simulate the low-resolution input $g_{MS}$ and $g_{PAN}$, while the original MS images were employed as the ground truth $f_{MS}$ to assess the qualities of the pan-sharpened results. Five numeric metrics were applied to quantify the qualities of the pan-sharpened images from the simulated experiments: the peak signal-to-noise ratio (PSNR) [44], the universal image quality metric (Q) [45], the *Erreur Relative Globale Adimensionnelle de Synthèse* (ERGAS) [46], thespectral angle mapper (SAM) [47], and $Q2^n$: An expanded version of Q that adds spectral fidelity into consideration[48]. The results of the simulated experiments are listed in Table II and Table III, and in each comparison group, the best performance is marked in **bold**.

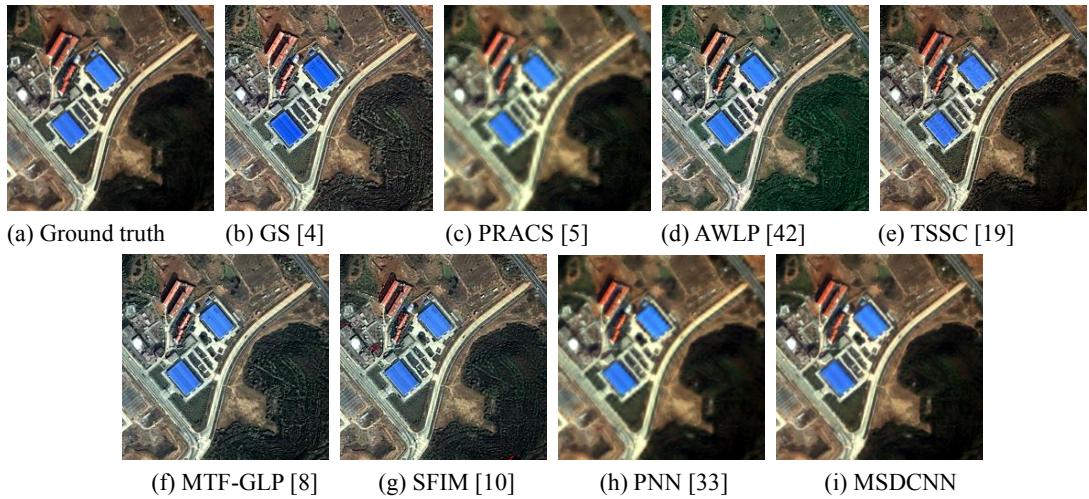

(a) Ground truth  (b) GS [4]  (c) PRACS [5]  (d) AWLP [42]  (e) TSSC [19]
(f) MTF-GLP [8]  (g) SFIM [10]  (h) PNN [33]  (i) MSDCNN

Fig. 7. Results of the simulated experiment on an area of industrial land, which was extracted from a QuickBird image of Yichang, China, obtained in 2015.

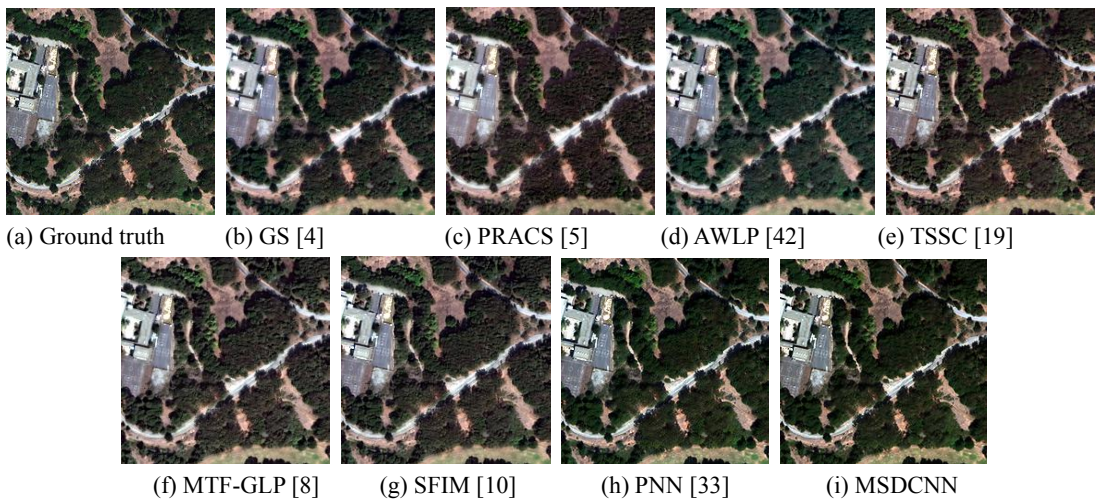

(a) Ground truth  (b) GS [4]  (c) PRACS [5]  (d) AWLP [42]  (e) TSSC [19]
(f) MTF-GLP [8]  (g) SFIM [10]  (h) PNN [33]  (i) MSDCNN

Fig. 8. Results of the simulated experiment on an area of city vegetation, which was extracted from a WorldView-2 image of San Francisco, United States, obtained in 2011.

TABLE II
NUMERIC ASSESSMENT OF THE SIMULATED
QUICKBIRD IMAGE PAN-SHARPENING.

| Bands | Algorithm | PSNR (↑) | Q (↑) | ERGAS (↓) | SAM (↓) | Q4 (↑) |
|---|---|---|---|---|---|---|
| 4 | GS [4] | 34.0907 | 0.8305 | 4.5014 | 4.0227 | 0.6831 |
| | PRACS [5] | 35.9282 | 0.8397 | 3.7501 | 3.5646 | 0.6138 |
| | MTF-GLP[8] | 34.3894 | 0.8227 | 4.4409 | 3.7893 | 0.6803 |
| | SFIM [10] | 34.4410 | 0.8264 | 5.1491 | 3.7708 | 0.6818 |
| | AWLP [42] | 34.2055 | 0.8314 | 4.0463 | 3.6587 | 0.6466 |
| | TSSC [19] | 35.3860 | 0.8488 | 3.9773 | 3.7154 | 0.7039 |
| | PNN [33] | 38.5201 | 0.9206 | 2.7110 | 2.6405 | 0.7569 |
| | MSDCNN | **39.2674** | **0.9303** | **2.5408** | **2.4605** | **0.7924** |

TABLE III
NUMERIC ASSESSMENT OF THE SIMULATED
WORLDVIEW-2 IMAGE PAN-SHARPENING.

| Bands | Algorithm | PSNR (↑) | Q (↑) | ERGAS (↓) | SAM (↓) | Q8 (↑) |
|---|---|---|---|---|---|---|
| 8 | GS [4] | 33.6506 | 0.8606 | 4.8395 | 6.1412 | 0.5781 |
| | PRACS [5] | 35.7979 | 0.8631 | 4.5579 | 6.2920 | 0.6849 |
| | MTF-GLP[8] | 34.8187 | 0.8788 | 4.3748 | 5.7698 | 0.6324 |
| | SFIM [10] | 34.8078 | 0.8756 | 4.3230 | 5.7579 | 0.6284 |
| | AWLP [42] | 35.0906 | 0.8769 | 4.4214 | 5.9263 | 0.6870 |
| | TSSC [19] | 36.7291 | 0.8951 | 3.9735 | 5.8269 | 0.6941 |
| | PNN [33] | 37.7634 | 0.9389 | 3.0695 | 4.4757 | 0.7697 |
| | MSDCNN | **38.1045** | **0.9570** | **2.9331** | **4.2483** | **0.7740** |

From the numeric assessment results listed above, the superiority of the two CNN-based algorithms compared to the traditional methods is clear, as under all the full-reference metrics, the performances of PNN and MSDCNN are far ahead of the other algorithms, while the lead status is held by MSDCNN. For the 240 tested image patches containing various ground objects, the impressive performance gains of the proposed network helps us to confirm that the multi-scale convolutional layer blocks significantly contribute to improving the robustness of the feature extraction and merging in all the bands along the spectral dimension.

As numeric metrics are applied to assess the quality of fused images from a quantifiable perspective, careful visual inspection is also needed to identify artifacts and distortions that elude the quantitative analysis. From the results of the simulated experiments, two groups of images that typically highlight the advantages and drawbacks of the various methods are selected and displayed in Figs. 7–8. For the purpose of displaying true-color images, the spectral bands covering the wavelengths of red, blue, and green light are selected according to the MS band division of the sensor, i.e., the 3rd, 2nd, and 1st bands of QuickBird, and the 5th, 3rd, and 2nd bands of WorldView-2.

By comparing the images displayed in Fig. 7 and Fig. 8, it can be seen that the results of the CNN-based methods are the most similar to the ground truth, both in spatial detail and spectral fidelity. For example, the vegetation areas in the lower-right of the group of images listed in Fig. 7. In particular, the proposed MSDCNN performs better than PNN [33] in preserving edges and the spectral features of ground objects with very small sizes, such as the concrete area in the middle-left of Fig. 7(h)-(i) and the bare soil in the upper-middle of Fig. 8(h)-(i). In some of the other six methods, while the spatial details are impressively sharpened and highlighted, noticeable spectral distortion is also apparent (GS [4], AWLP [42], MTF-GLP [8], and SFIM [10]). In contrast, better colors are obtained in the results of PRACS [5], but the restoration of spatial information is still not satisfactory. Fig. 7(e) shows that for the QuickBird dataset, TSSC [19] is a well-balanced solution, but when it came to the WorldView-2 dataset, there is still a gap between the performance of the sparse representation based model and the proposed MSDCNN.

The comparisons strongly support our statement that for remote sensing images with multiple sources that do not fully overlap in the spectral domain, non-linear models based on deep learning are better able to handle the fusion task. It should also be noted that compared to the related PAN image and some of the over-sharpened fusion results, the slightly "blurry" appearance is also shared by the ground truth and the result of MSDCNN, which indicates that instead of being constrained by artificially given priors, the proposed network is able to fit various types of transformation.

*C. Real-Data Experiments*

Original MS and PAN images were also input into the models to yield full-resolution results. There are non-reference numeric metrics that can quantify the qualities of pan-sharpened images, i.e., the quality with no-reference index (QNR) [49] and the spatial and spectral components of it ($D_S$ and $D_\lambda$). We employed the three metrics for the quantitative assessment of the real-data experiments, and the results are listed Table IV.

However, considering that these metrics are computed with $G_{MS}$ and $g_{PAN}$ as references, instead of the unattainable ground truth, we should note that what can be quantified by such metrics is the similarity of certain components in the fused images to the low-resolution observations, but not the real fidelity at the level of high resolution. The comparisons in Table IV also support our assumption, as the results of PRACS [5] are very similar to the related low-resolution MS images and barely sharpened in the spatial domain, but by the similarity, they achieved very high $D_\lambda$ values and jointly improved their QNR index to a state-of-the-art level.

Thus, in the following discussion, the real-data experiments are mainly discussed based on the visual inspection, instead of the three numeric metrics. Three ground regions were selected from the pan-sharpened full-resolution images to be investigated, as displayed in Figs. 9–11.

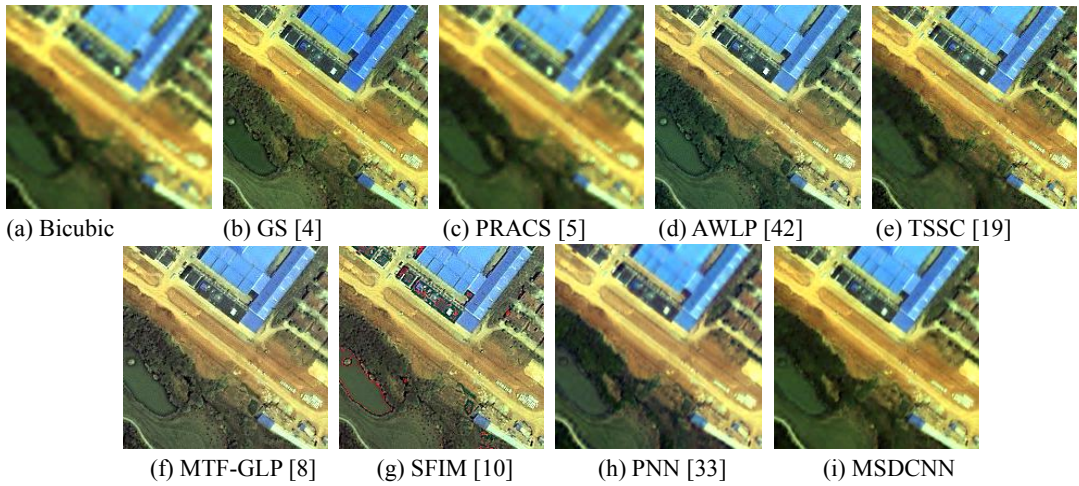

Fig. 9. Results of the real-data experiment on an area of industrial land, which was extracted from an IKONOS image of Wuhan, China.

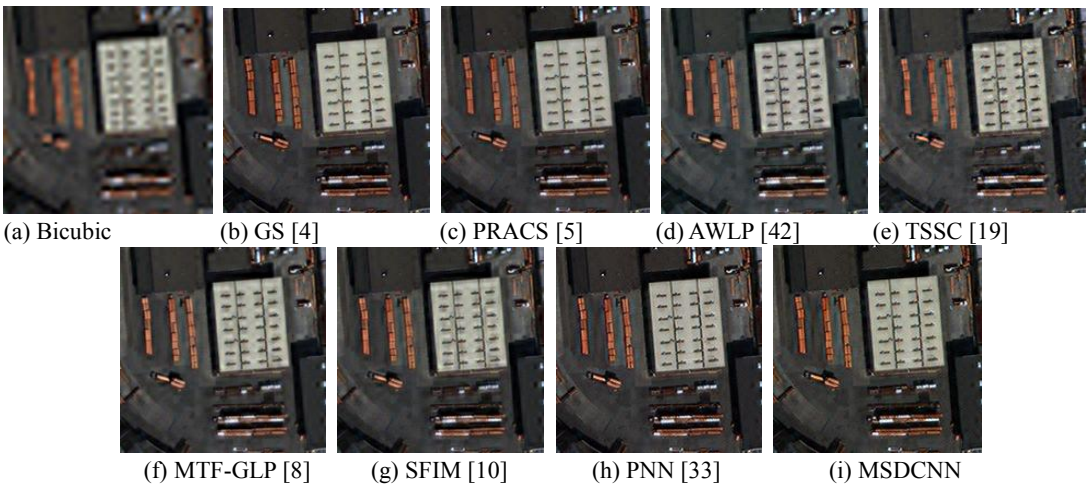

Fig. 10. Results of the real-data experiment on an area of impervious surface, which was extracted from a WorldView-2 image of San Francisco, United States, obtained in 2011.

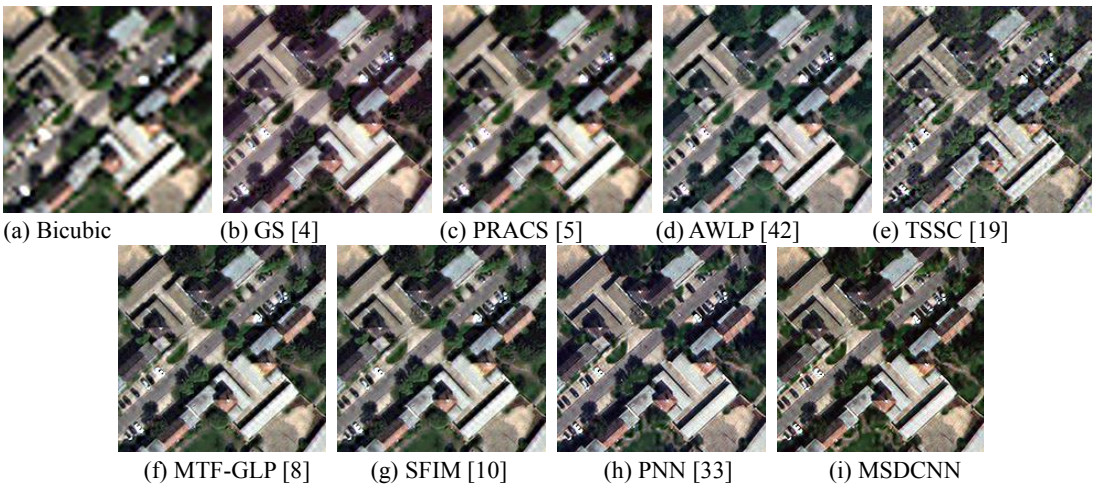

Fig. 11. Results of the real-data experiment on an area of urban vegetation, which was extracted from a WorldView-2 image of San Francisco, United States, obtained in 2011.

By comparing the images displayed in Fig. 9, we can observe a tendency that is similar to the story told by the previous simulated experiments: MSDCNN and PNN [33] return images with the best spectral fidelity and appropriately sharpened spatial details, while the proposed network performs slightly better in preserving details with small sizes. Among the other compared methods, TSSC [19] remains competitive in the real-data experiments, which is supported by the high quality of Fig. 9(e) and its high similarity to the related image obtained by MSDCNN in Fig. 9(i). However, when it comes to the WorldView-2 dataset, as shown in Fig. 10(e) and Fig. 11(e), the performance of TSSC becomes less robust, while MSDCNN is still able to avoid introducing ringing artifacts from the up-sampled MS images and prevents spectral distortion, for example, the impressive quality of Fig. 11(i) shows that though the MS image in Fig. 11(a) is severely corrupted after interpolation, our proposed network still performed a good fusion with the guidance from its related PAN image.

TABLE IV
NUMERIC ASSESSMENT OF REAL-DATA IKONOS AND WORLDVIEW-2 IMAGE PAN-SHARPENING.

| Bands | Algorithm | QNR (↑) | $D_S$ (↓) | $D_\lambda$ (↓) |
|---|---|---|---|---|
| | | IKONOS | | |
| 4 | GS [4] | 0.7661 | 0.1753 | 0.0729 |
| | PRACS [5] | 0.8451 | 0.1183 | 0.0445 |
| | MTF-GLP [8] | 0.7434 | 0.1580 | 0.1202 |
| | SFIM [10] | 0.7526 | 0.1601 | 0.1068 |
| | AWLP [42] | 0.7433 | 0.1634 | 0.1148 |
| | TSSC [19] | 0.8587 | 0.0997 | 0.0497 |
| | PNN [33] | 0.8606 | 0.0895 | 0.0555 |
| | MSDCNN | **0.8797** | **0.0774** | **0.0469** |
| | | WorldView-2 | | |
| 8 | GS [4] | 0.8403 | 0.1264 | 0.0415 |
| | PRACS [5] | **0.8916** | 0.0892 | **0.0224** |
| | MTF-GLP [8] | 0.8208 | 0.1108 | 0.0797 |
| | SFIM [10] | 0.8380 | 0.1073 | 0.0645 |
| | AWLP [42] | 0.8458 | 0.0991 | 0.0635 |
| | TSSC [19] | 0.8425 | 0.1037 | 0.0617 |
| | PNN [33] | 0.8725 | 0.0826 | 0.0538 |
| | MSDCNN | 0.8893 | **0.0779** | 0.0390 |

*D. Further Discussion*

In this sub-section, the default settings of MSDCNN used in the experiments are compared with the alternatives. The performance of the network with different settings was tested by simulated experiments on the QuickBird dataset containing 160 images and assessed with the full-reference Q and ERGAS metrics.

**1) Setting Hyper-Parameters for Training MSDCNN:** As mentioned above, the momentum and learning rate are initialized as $\mu = 0.9$ and $\varepsilon = 0.1$, and for every 60 epochs, $\varepsilon$ is multiplied by $\gamma = 0.5$, while $\mu$ is fixed as 0.9. From the performance-to-epoch curves in Fig. 12, we can see that the residual learning architecture of MSDCNN helps the network to quickly reach state-of-the-art accuracy within about 50 training epochs, while the ceiling of its performance is still far away. Although the curves in Fig. 12 indicate that the default settings work well, we tried another two settings for the learning rate $\gamma$ to confirm our understanding of the learning process. The results of the comparison are shown in Fig. 13.

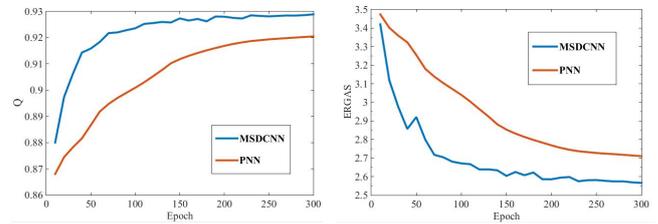

Fig. 12. Average Q and ERGAS of MSDCNN and PNN on the QuickBird dataset.

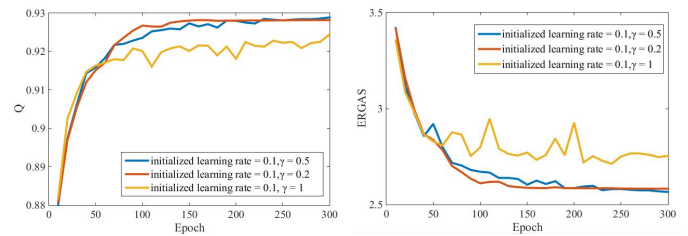

Fig. 13. Average Q and ERGAS of MSDCNN with different values of $\gamma$.

Fig. 13 helps us to confirm that the default setting of $\gamma = 0.5$ is a balanced decision between error decrease in the early training epochs and relatively smooth convergence in the later stages. Meanwhile, setting an appropriately low value for $\gamma$ can lead to earlier convergence, but when $\gamma$ is too small, the opportunity of breaking out of local minima may be lost.

**2) Connection Architecture of the Multi-Scale Convolutional Layer Blocks:** In the default architecture of MSDCNN, there is a flat convolutional layer between two multi-scale blocks to reduce the spectral dimension from 60 to 30. To confirm its validity, two different architectures were compared, and their connections are illustrated in Fig. 14. In Block 2, two multi-scale layers are contained in each block. In Block 3, a further skip connection is used, as in [27], and thus the spectral dimensionality is kept until the image is fed into the last layer.

By comparing the curves shown in Fig. 15, we can confirm that reducing the spectral dimensionality is necessary for the task, as the effect of using Block 3 without the reduction layer appears to be negative. From the comparison between Block 1 and Block 2, we can observe that the deeper architecture needs more training epochs to reach a convergence region with a slightly higher accuracy, but such limited improvement is still far away from our expectation, and we assume that the network formed by Block 2 is not deep enough to fully develop the advantages of residual learning. Possible ways to reduce the training time cost will be studied in our future work.

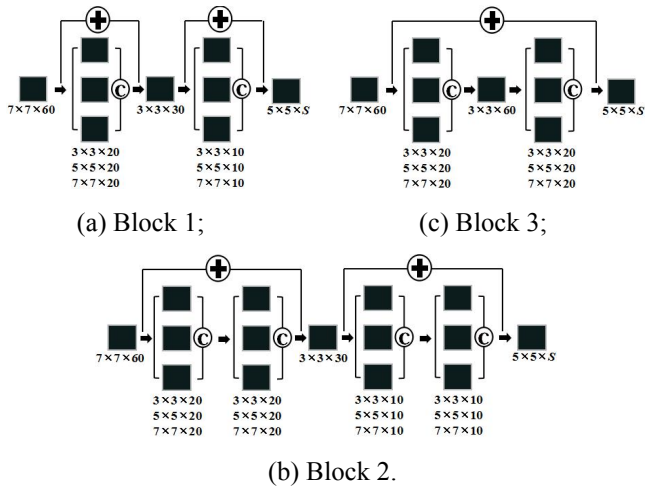

(a) Block 1;   (c) Block 3;

(b) Block 2.

Fig. 14. Block 1 is the architecture used in all the experiments undertaken in this study.

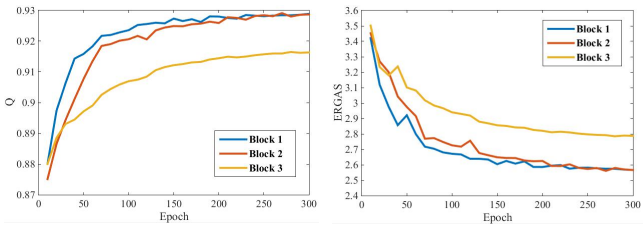

Fig. 15. Average Q and ERGAS of MSDCNN with different values of $\gamma$ on the QuickBird dataset.

## V. CONCLUSION

In this paper, we have proposed a new CNN architecture for remote sensing imagery pan-sharpening. The main innovations in the model are the concepts of multi-scale extraction, multi-depth sharing, and merging of features from the spatial domain of the MS and PAN images. Compared to many of the traditional and state-of-the-art pan-sharpening algorithms, the results of experiments undertaken on different datasets strongly indicate that the proposed MSDCNN is able to yield high-quality images with the best quantitative fidelity and appropriate sharpness.

In our future work, as the art of designing CNNs has not yet been fully explained from an analytical perspective instead of empirical ideas, there is still scope for the architecture of the proposed network to be optimized. Furthermore, our current feature learning strategies also require much study to transfer the obtained knowledge to some extended fields of remote sensing image fusion, quality improvement, and interpretation tasks, such as spatial-temporal unified fusion[50], hyperspectral image denoising[51][52], aerial scene classification[53][54] and target detection[55]. To satisfy the needs of the topics listed above, we also expect to develop advanced techniques of network compression and training data generalization, which makes help to effectively process routine tasks on an application level.